\documentclass[runningheads]{llncs}

\usepackage[T1]{fontenc}
\usepackage{graphicx}
\usepackage{amsmath}
\usepackage{amsfonts}
\usepackage{url}
\usepackage{float}
\usepackage{tikz}
\usetikzlibrary{positioning, arrows.meta, fit, backgrounds}
\usepackage{booktabs}

\begin{document}
\title{DR-Mamba: Automatic Inference-Time Domain Adaptation for Document Image Binarization via Sample-Conditioned Detail-Background Suppression}
\titlerunning{DR-Mamba for Document Image Binarization}
\author{Sheng-Wei Chan\inst{1} \and Jen-Shiun Chiang\inst{1}}
\authorrunning{S.-W. Chan et al.}
\institute{
Department of Electrical and Computer Engineering, Tamkang University, New Taipei City, Taiwan \\
\email{412440330@o365.tku.edu.tw, chiang@mail.tku.edu.tw}
}
\maketitle

\begin{abstract}
Degraded document image binarization is sensitive to domain shifts caused by paper aging, bleed-through, stains, shadows, and uneven illumination, and the foreground-background separation of recent learning-based methods can become unstable on unseen degradation domains. We propose DR-Mamba, a sample-conditioned detail-background suppression framework that performs automatic inference-time domain adaptation for document image binarization. Unlike test-time adaptation methods that require gradient updates or auxiliary data at inference, DR-Mamba adapts to each input document through input-dependent gates within a single forward pass, requiring no target-domain labels, no fine-tuning, and no test-time parameter updates. Instead of using Mamba-style selective scanning as a single generic feature path, DR-Mamba reinterprets it as fast--slow route modeling: a fast detail route captures local stroke structures, while a slow background route accumulates spatially persistent degradation responses. The two routes are integrated through an input-dependent subtractive gate, $D-\beta B$, which explicitly suppresses background interference rather than fusing features by addition or concatenation. We further add full-resolution detail-guided reconstruction and thin-stroke-aware supervision to recover fine strokes lost during downsampling. Evaluated under a leave-one-year-out protocol on DIBCO-style benchmarks, where each held-out year is treated as an unseen degradation domain, DR-Mamba shows that per-document, per-location subtractive suppression improves cross-domain robustness, with particularly strong performance on the most severely degraded held-out fold.

\keywords{Document image binarization \and Inference-time domain adaptation \and Automatic domain-adapted document analysis \and Sample-conditioned suppression \and Dual-Route Mamba}
\end{abstract}

\section{Introduction}
Document image binarization separates foreground strokes from background regions and remains a fundamental step in document image analysis~\cite{otsu1979threshold,sauvola2000adaptive,gatos2006adaptive}. Although learning-based methods have improved binarization quality~\cite{tensmeyer2017fcn,he2019deepotsu,deng2020document,zhao2019cascaded,souibgui2022docentr}, degraded documents still exhibit substantial domain shifts caused by acquisition devices, paper aging, ink diffusion, bleed-through, stains, shadows, and uneven illumination. These shifts are reflected in DIBCO and H-DIBCO benchmarks, where different years contain different document types and degradation patterns~\cite{gatos2009dibco,pratikakis2010hdibco,pratikakis2013dibco,pratikakis2017dibco,pratikakis2018hdibco}, and they reappear whenever a model is applied to new archives, scanners, writing materials, or acquisition environments. Since collecting pixel-level labels for every new collection is costly and often impractical, robust binarization should not rely on target-domain annotation or fine-tuning, but should automatically adjust to the degradation characteristics of each input sample. This motivates us to study degraded document binarization as an automatic, inference-time domain adaptation problem, in which DR-Mamba adapts to each input document through input-dependent gates within a single forward pass, requiring no target-domain labels, no fine-tuning, and no test-time parameter updates. This forward-pass, label-free form of per-sample adaptation falls directly under the scope of \emph{visual-specific automatic adaptation} for document analysis: each new document acts as its own implicit adaptation signal. We evaluate this setting with a leave-one-year-out protocol where each held-out benchmark year is treated as an unseen degradation domain, which is more stringent than random image-level splitting, as no sample from the target year is seen during training. A key challenge in this setting is foreground-background entanglement: text strokes and background artifacts may share similar local edges, textures, or low-contrast patterns. Simply applying a stronger backbone or a single long-range modeling path may therefore still mix foreground details with background degradation. Recent state space and Mamba-based visual models offer efficient long-range sequence modeling~\cite{gu2021s4,gu2023mamba,zhu2024visionmamba,liu2024vmamba}, but a single selective scanning route is not tailored to the suppression nature of document binarization. We propose DR-Mamba, which reformulates selective scanning as fast--slow route modeling rather than a generic feature operation. The fast detail route responds to local stroke structures, while the slow background route accumulates spatially persistent degradation responses, and the two are combined by an input-dependent subtractive gate, $D-\beta B$, that explicitly suppresses background interference instead of fusing all responses by addition or concatenation. A full-resolution detail-guided reconstruction module further restores high-frequency stroke cues lost during downsampling. Our contributions are summarized as follows:
\begin{itemize}
    \setlength{\itemsep}{0pt}\setlength{\parskip}{0pt}
    \item We formulate degraded document image binarization as an \emph{automatic, inference-time domain adaptation} problem, evaluated with a leave-one-year-out protocol where each held-out DIBCO/H-DIBCO year is treated as an unseen degradation domain.
    \item We propose a task-oriented Dual-Route Mamba block that reinterprets selective scanning as fast--slow detail-background modeling for degraded document binarization.
    \item We introduce an adaptive subtractive suppression mechanism, $D-\beta B$, whose suppression strength is predicted per input sample and per location. The gate $\beta$ thus serves as a forward-pass adaptation signal that automatically reconfigures the model's background-removal behavior for each unseen document, realizing visual-specific automatic domain adaptation without any target-domain label, gradient update, or fine-tuning step.
    \item We combine full-resolution detail-guided reconstruction with thin-stroke-aware supervision, and verify the contribution of each component through controlled ablations.
\end{itemize}

\section{Proposed Method}
 
\subsection{Overview}
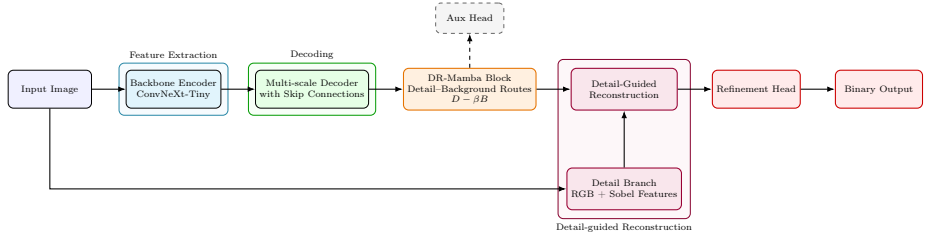
\begin{figure}[t]
\centering
\resizebox{\textwidth}{!}{
\begin{tikzpicture}[
    font=\scriptsize,
    node distance=0.8cm and 0.9cm,
    block/.style={
        draw, rounded corners, align=center,
        minimum height=1.0cm, minimum width=2.2cm, fill=blue!6
    },
    enc/.style={
        draw, rounded corners, align=center,
        minimum height=1.0cm, minimum width=2.3cm, fill=cyan!10
    },
    dec/.style={
        draw, rounded corners, align=center,
        minimum height=1.0cm, minimum width=2.4cm, fill=green!10
    },
    mamba/.style={
        draw=orange!90!black, thick, rounded corners, align=center,
        minimum height=1.1cm, minimum width=2.7cm, fill=orange!12
    },
    fusion/.style={
        draw=purple!80!black, thick, rounded corners, align=center,
        minimum height=1.1cm, minimum width=2.8cm, fill=purple!10
    },
    head/.style={
        draw=red!80!black, thick, rounded corners, align=center,
        minimum height=1.0cm, minimum width=2.3cm, fill=red!8
    },
    aux/.style={
        draw=gray!80!black, dashed, rounded corners, align=center,
        minimum height=0.8cm, minimum width=1.8cm, fill=gray!8
    },
    arrow/.style={-{Latex[length=2mm]}, thick},
    skip/.style={-{Latex[length=2mm]}, thick, dashed}
]
 
\node[block] (input) {Input Image};
\node[enc, right=of input] (encoder) {Backbone Encoder\\ConvNeXt-Tiny};
\node[dec, right=of encoder] (decoder) {Multi-scale Decoder\\with Skip Connections};
\node[mamba, right=of decoder] (mamba) {DR-Mamba Block\\Detail--Background Routes\\$D-\beta B$};
\node[fusion, right=of mamba] (fusion) {Detail-Guided\\Reconstruction};
\node[head, right=of fusion] (refine) {Refinement Head};
\node[head, right=of refine] (output) {Binary Output};
 
\node[fusion, below=1.5cm of fusion] (detail) {Detail Branch\\RGB + Sobel Features};
 
\node[aux, above=0.9cm of mamba] (auxhead) {Aux Head};
 
\draw[arrow] (input) -- (encoder);
\draw[arrow] (encoder) -- (decoder);
\draw[arrow] (decoder) -- (mamba);
\draw[arrow] (mamba) -- (fusion);
\draw[arrow] (fusion) -- (refine);
\draw[arrow] (refine) -- (output);
 
\draw[arrow] (input) |- (detail);
\draw[arrow] (detail) -- (fusion);
 
\draw[skip] (mamba) -- (auxhead);
 
\begin{scope}[on background layer]
\node[draw=cyan!60!black, rounded corners, fill=cyan!3, fit=(encoder),
      inner sep=0.18cm, label={[font=\scriptsize]above:Feature Extraction}] {};
\node[draw=green!60!black, rounded corners, fill=green!3, fit=(decoder),
      inner sep=0.18cm, label={[font=\scriptsize]above:Decoding}] {};
\node[draw=purple!70!black, rounded corners, fill=purple!3, fit=(detail)(fusion),
      inner sep=0.22cm, label={[font=\scriptsize]below:Detail-guided Reconstruction}] {};
\end{scope}
 
\end{tikzpicture}
}
\caption{Simplified overview of the proposed DR-Mamba framework. A ConvNeXt-Tiny backbone extracts hierarchical document features, which are restored by a multi-scale decoder. The proposed DR-Mamba block models the decoded representation through a fast detail route and a slow background route, followed by adaptive subtractive suppression to reduce degradation interference. Full-resolution edge cues are further introduced through detail-guided reconstruction to improve stroke continuity and boundary sharpness.}
\label{fig:dr_mamba_overview}
\end{figure}

The proposed DR-Mamba framework targets cross-domain degraded document image binarization. Given an input image $I \in \mathbb{R}^{3 \times H \times W}$, the goal is to predict a foreground probability map $\hat{Y} \in [0,1]^{H \times W}$ and obtain the final binary result by thresholding at $0.5$. As shown in Fig.~\ref{fig:dr_mamba_overview}, the framework contains four components: a ConvNeXt-Tiny encoder, a multi-scale decoder, the proposed DR-Mamba block, and a detail-guided reconstruction module. The design is motivated by viewing degraded binarization as a background suppression problem in addition to foreground detection. DR-Mamba addresses this by separating the decoded representation into a fast detail route and a slow background route, fused through adaptive subtraction instead of generic addition or concatenation. Crucially, the strength of this suppression is not a fixed hyper-parameter but is predicted per sample and per location from the input itself, realizing the automatic inference-time adaptation described above.

\subsection{Backbone Encoder and Multi-Scale Decoder}
We adopt a standard hierarchical encoder-decoder backbone for dense prediction; it is not a contribution of this work, so we describe it only briefly. The encoder extracts four multi-level features from the input image,
\begin{equation}
    \{F_1, F_2, F_3, F_4\} = E(I),
\end{equation}
where $E(\cdot)$ is a ConvNeXt-Tiny encoder~\cite{liu2022convnext} and $F_i$ is the feature map at stride $2^{i+1}$ (i.e.\ $F_1$ is the shallowest/highest-resolution map and $F_4$ the deepest/lowest-resolution one).
 
The decoder restores spatial resolution by fusing high-level semantic features with low-level spatial features through skip connections. Starting from $F_4$, it applies a sequence of upsample-and-fuse operations:
\begin{align}
    U_4 &= \mathcal{D}_4\!\left(\mathrm{Up}(F_4),\, F_3\right), \\
    U_3 &= \mathcal{D}_3\!\left(\mathrm{Up}(U_4),\, F_2\right), \\
    U_2 &= \mathcal{D}_2\!\left(\mathrm{Up}(U_3),\, F_1\right).
\end{align}
Here $\mathrm{Up}(\cdot)$ denotes a $\times 2$ upsampling operator, and each decoder block $\mathcal{D}_i(\cdot)$ concatenates its upsampled input with the corresponding skip feature and applies a convolution--normalization--activation fusion. The intermediate features $U_4$, $U_3$, and $U_2$ are therefore the decoder outputs at successively higher resolutions, with the level index $i$ matching the skip feature $F_i$ being fused. The final decoded feature $U_2$ carries both semantic context and spatial detail at stride~$4$, and is passed to the proposed DR-Mamba block for detail--background decomposition.
 
\subsection{Dual-Route Mamba Block}
 
The DR-Mamba block is the core of our framework. It adapts Mamba-style selective scanning~\cite{gu2023mamba} to the \emph{suppression} nature of degraded document binarization. Throughout this section, $\odot$ denotes the element-wise (Hadamard) product, $\sigma(\cdot)$ the sigmoid function, and $\mathrm{softplus}(x)=\log(1+e^{x})$ a smooth positive map.
 
\paragraph{Input projection and per-token gates.}
Given the decoded feature $U_2$, we first apply an input projection
\begin{equation}
    Z = \phi(U_2),
\end{equation}
where $\phi(\cdot)$ is the composition of layer normalization, a linear projection, a $3\times3$ depth-wise convolution, and a SiLU activation; $Z\in\mathbb{R}^{(H_2 W_2)\times d}$ is the resulting per-token feature, $H_2\times W_2$ is the spatial size of $U_2$, $d$ is the block width, and $t\in\{1,\dots,H_2 W_2\}$ indexes a spatial token with feature $z_t\in\mathbb{R}^{d}$. From $Z$, the block predicts four per-token quantities,
\begin{equation}
    s_t=\sigma(W_s z_t),\quad
    g_t=\sigma(W_g z_t),\quad
    \Delta_t=\mathrm{softplus}(W_{\Delta}z_t),\quad
    \beta_t=\sigma(W_{\beta}z_t),
\end{equation}
where $W_s, W_g, W_{\Delta}, W_{\beta}\in\mathbb{R}^{d\times d}$ are learnable projection matrices. Here $s_t\in(0,1)^d$ is a \emph{structure gate} that decides how much of each channel is routed to the detail route versus the background route; $g_t\in(0,1)^d$ is an \emph{input modulation gate} controlling how much new evidence enters the state at token $t$; $\Delta_t\in\mathbb{R}_{>0}^d$ is an input-dependent time scale (the selective $\Delta$ of Mamba); and $\beta_t\in(0,1)^d$ is the \emph{suppression gate}. Because $\beta_t$ is spatial- and channel-dependent, the model adjusts the amount of background suppression according to each input document and each local degradation pattern; the gate $\beta_t$ thus implements per-sample, per-location adaptation entirely within the forward pass, with no target-domain supervision and no parameter update required.
 
\paragraph{Fast--slow selective dynamics.}
Instead of forming two routes by channel splitting, DR-Mamba assigns the two routes \emph{different} memory horizons. We define two positive per-channel decay rates $A_D, A_B\in\mathbb{R}_{>0}^{d}$ for the detail and background routes:
\begin{equation}
    A_D = A_B + \mathrm{softplus}(A_{\mathrm{gap}}),\qquad A_D > A_B .
\end{equation}
Here $A_B$ is a learnable per-channel decay rate (the slow route), and $A_{\mathrm{gap}}\in\mathbb{R}^{d}$ is a learnable per-channel parameter that parameterizes the \emph{gap} between the two routes. Passing $A_{\mathrm{gap}}$ through $\mathrm{softplus}$ guarantees a non-negative gap, so the constraint $A_D>A_B$ holds \emph{by construction} rather than being imposed by clipping or an external penalty. A larger decay rate means faster forgetting: the detail route ($A_D$) responds quickly to local high-frequency stroke changes, while the background route ($A_B$) integrates spatially persistent degradation responses such as stains and bleed-through. The corresponding input-dependent decay factors at token $t$ are
\begin{equation}
    a^D_t = \exp(-\Delta_t\, A_D),\qquad
    a^B_t = \exp(-\Delta_t\, A_B),
\end{equation}
where $a^D_t, a^B_t\in(0,1)^d$ vary with both content (through $\Delta_t$) and channel (through $A_D, A_B$), so the fast/slow split is selective rather than a fixed constant.
 
\paragraph{Routes and adaptive subtractive suppression.}
The two selective states are updated by the recurrences
\begin{align}
    D_t &= a^D_t \odot D_{t-1} + g_t \odot s_t, \\
    B_t &= a^B_t \odot B_{t-1} + g_t \odot (1-s_t),
\end{align}
where $D_t, B_t\in\mathbb{R}^{d}$ are the detail and background route states. The gated input $g_t\odot s_t$ writes detail-sensitive evidence into the fast route, while $g_t\odot(1-s_t)$ writes the complementary, degradation-prone evidence into the slow route. The two routes are then fused by adaptive subtractive suppression,
\begin{equation}
    O_t = D_t - \beta_t \odot B_t ,
\end{equation}
where $O_t\in\mathbb{R}^{d}$ is the per-token output of the route fusion. This differs fundamentally from generic feature fusion: addition or concatenation preserves \emph{both} foreground and degradation responses, whereas binarization requires the background response to be \emph{removed}. The subtractive form injects this task-specific prior directly into the operator, and the input-dependent gate $\beta_t$ controls how aggressively the accumulated background state is cancelled at each location. In our ablations this sample-conditioned subtraction is the single largest source of gain (Sec.~\ref{sec:ablation}), which is why we keep it at the center of the design.
 
\paragraph{Four-directional scan.}
A causal recurrence only models dependencies along one traversal order, but document strokes extend in arbitrary 2-D directions. We therefore aggregate the recurrence over multiple scan orders,
\begin{equation}
    O_{\mathrm{DR}} = \sum_{r \in \mathcal{R}} \mathrm{Scan}_r(D, B, \beta),
\end{equation}
where each $\mathrm{Scan}_r$ realizes the route recurrences and subtractive fusion above along one traversal order $r$: the 2-D feature map is first \emph{serialized} into a 1-D token sequence following $r$, the selective recurrence is computed in linear time along that sequence to produce $\{O_t\}$, and the outputs are then \emph{scattered back} to their original spatial positions. The set $\mathcal{R}=\{\text{left}\!\to\!\text{right},\ \text{right}\!\to\!\text{left},\ \text{top}\!\to\!\text{bottom},\ \text{bottom}\!\to\!\text{top}\}$ contains the four axis-aligned orders, and $O_{\mathrm{DR}}$ sums their results so that horizontal and vertical stroke continuity are both captured. This four-directional design is sufficient to cover dominant stroke orientations while avoiding the over-smoothing and extra computation we observe with denser directional aggregation (Sec.~\ref{sec:ablation}).
 
\subsection{Detail-Guided Reconstruction}
 
Encoder-decoder networks lose fine spatial detail through downsampling, which is especially harmful for thin strokes and sharp boundaries. To compensate, we add a lightweight detail branch that operates at full resolution. We obtain horizontal and vertical edge responses with fixed Sobel filters,
\begin{equation}
    G_x = \mathrm{Sobel}_x(I), \qquad
    G_y = \mathrm{Sobel}_y(I),
\end{equation}
where $\mathrm{Sobel}_x(\cdot)$ and $\mathrm{Sobel}_y(\cdot)$ apply the standard $3\times3$ horizontal and vertical Sobel kernels as fixed, non-learnable depth-wise convolutions on $I$, and $G_x, G_y$ are the resulting full-resolution gradient maps. The original image and the magnitudes of the two edge maps are concatenated and passed through a small convolutional branch:
\begin{equation}
    F_{\mathrm{detail}} = \psi\!\left([\,I,\ |G_x|,\ |G_y|\,]\right),
\end{equation}
where $[\cdot]$ denotes channel-wise concatenation, $|\cdot|$ the element-wise absolute value, $\psi(\cdot)$ a two-layer convolutional branch, and $F_{\mathrm{detail}}$ the full-resolution high-frequency feature.
 
The output of the DR-Mamba block is upsampled to the input resolution and fused with $F_{\mathrm{detail}}$:
\begin{equation}
    F_{\mathrm{fuse}} = \mathcal{F}\!\left([\,\mathrm{Up}(O_{\mathrm{DR}}),\ F_{\mathrm{detail}}\,]\right),
\end{equation}
where $\mathrm{Up}(\cdot)$ here is a learned sub-pixel upsampling, $\mathcal{F}(\cdot)$ is a convolutional fusion operation, and $F_{\mathrm{fuse}}$ is the fused feature that combines global context from DR-Mamba with local high-frequency cues from the original image. A refinement head then produces the binarization logit:
\begin{equation}
    P = \mathcal{H}(F_{\mathrm{fuse}}),
\end{equation}
where $\mathcal{H}(\cdot)$ is a refinement-and-prediction head built from residual convolutional blocks with different dilation rates (to enlarge the receptive field at full resolution) followed by a $1\times1$ prediction layer, and $P\in\mathbb{R}^{H\times W}$ is the predicted logit map. The final foreground probability map is
\begin{equation}
    \hat{Y} = \sigma(P),
\end{equation}
i.e.\ the pixel-wise sigmoid of the logit map; the binary output is obtained by thresholding $\hat{Y}$ at $0.5$.
 
\subsection{Training Objective}
 
Document binarization is sensitive to thin strokes, boundary errors, and foreground--background imbalance, so we train with a thin-stroke-aware compound loss rather than binary cross-entropy alone. The main loss has six terms:
\begin{equation}
\begin{aligned}
    \mathcal{L}_{main}
    = &\,\lambda_{sdf}\mathcal{L}_{sdf}
    + \lambda_{bce}\mathcal{L}_{bce}
    + \lambda_{tv}\mathcal{L}_{tv}
    + \lambda_{pfm}\mathcal{L}_{pfm} \\
    &+ \lambda_{cldice}\mathcal{L}_{cldice}
    + \lambda_{bnd}\mathcal{L}_{bnd}.
\end{aligned}
\end{equation}
 
The dominant term $\mathcal{L}_{sdf}$ is a signed distance field loss that aligns the zero-level set of the predicted logit field with the stroke boundary. We convert the ground-truth mask $Y$ into a normalized signed distance map $S(Y)\in[-1,1]^{H\times W}$: for each pixel, $S(Y)$ is the signed Euclidean distance to the nearest stroke boundary --- positive inside strokes, negative in the background, and exactly zero on the boundary --- clipped to $[-T,T]$ and divided by $T$, where the normalization constant $T$ is the saturation distance (in pixels) beyond which the field is flat. The SDF loss is then
\begin{equation}
    \mathcal{L}_{sdf}
    = \mathrm{SmoothL1}\!\left(\tanh\!\left(\tfrac{P}{T}\right),\, S(Y)\right),
\end{equation}
where $\tanh(P/T)$ squashes the logit field into the same $[-1,1]$ range as the target. By forcing $\{P=0\}$ onto the stroke boundary, this term yields sharp, well-aligned edges and a \emph{threshold-free} decision ($P>0 \Leftrightarrow \hat{Y}>0.5$), which improves boundary localization and reduces DRD relative to a BCE baseline (Table~\ref{tab:ablation_loss}).
 
The remaining terms are standard and act as light, complementary regularizers. $\mathcal{L}_{bce}$ is a pixel-wise binary cross-entropy anchor. $\mathcal{L}_{tv}$ is a Tversky loss~\cite{salehi2017tversky} that handles foreground--background imbalance by weighting false positives and false negatives asymmetrically; we use $\alpha=0.70$ and $\beta=0.30$ to discourage overly thick strokes. $\mathcal{L}_{pfm}$ is a soft pseudo-F-measure loss that improves thin-stroke recall on the skeletonized ground truth. $\mathcal{L}_{cldice}$ is a clDice loss~\cite{shit2021cldice} that preserves stroke connectivity, and $\mathcal{L}_{bnd}$ is a boundary penalty~\cite{kervadec2019boundary} that suppresses false positives in the dilated boundary region. The weights are
\begin{equation}
    \lambda_{sdf}=1.0,\quad
    \lambda_{bce}=0.1,\quad
    \lambda_{tv}=0.3,\quad
    \lambda_{pfm}=0.5,\quad
    \lambda_{cldice}=0.2,\quad
    \lambda_{bnd}=0.2,
\end{equation}
which are selected empirically on the validation split of the leave-one-year-out folds (Sec.~3.2). We deliberately keep $\lambda_{sdf}=1.0$ dominant, because the SDF term carries the threshold-free boundary signal, while the other terms are down-weighted so they regularize rather than compete with it; in particular $\lambda_{bce}$ is small because a strong BCE fights the bounded-linear field that the SDF term builds near the boundary.
 
Finally, an auxiliary prediction head is attached after the DR-Mamba block for deep supervision at a lower resolution. The auxiliary target $Y_{aux}$ is obtained by adaptive max pooling the ground-truth mask $Y$ to the auxiliary output resolution, and $\mathcal{L}_{aux}$ is a binary cross-entropy loss between the auxiliary prediction and $Y_{aux}$. The full training objective is
\begin{equation}
    \mathcal{L} = \mathcal{L}_{main} + \lambda_{aux}\mathcal{L}_{aux},\qquad \lambda_{aux}=0.3 .
\end{equation}
Such imbalance-aware, boundary-aware, and topology-aware losses have been widely used in dense prediction and thin-structure segmentation to improve class balance, boundary alignment, and connectivity preservation~\cite{salehi2017tversky,kervadec2019boundary,shit2021cldice}.

\section{Experiments}
\subsection{Experimental Setup}
All experiments are implemented in PyTorch~\cite{paszke2019pytorch} and conducted on a single NVIDIA RTX 5080 GPU with 16GB memory. The full model has 33.51M parameters and 178 GFLOPs at a $512\times512$ input. DR-Mamba uses a ConvNeXt-Tiny backbone~\cite{liu2022convnext} pretrained on ImageNet~\cite{deng2009imagenet} as the encoder. The encoder extracts four hierarchical feature maps with channel dimensions of 96, 192, 384, and 768, which are progressively fused by a U-shaped decoder before entering the proposed DR-Mamba block. During training, document images are randomly cropped into $512 \times 512$ patches. The batch size is set to 4, and gradient accumulation is applied for 4 steps. The model is optimized using AdamW~\cite{loshchilov2019adamw} and trained for 150 epochs with a maximum learning rate of $2 \times 10^{-4}$. We adopt bfloat16 mixed-precision training~\cite{micikevicius2018mixed} to reduce GPU memory consumption; we specifically use bfloat16 rather than fp16 because the selective scan operations in the Mamba block are numerically sensitive and prone to overflow under fp16. During testing, full-resolution document images are processed by sliding-window inference with a crop size of 512 and a stride of 256.

To expose the model to diverse degradation conditions, we apply degradation-oriented data augmentation during training: bleed-through-like interference, synthetic paper texture, ink-like stains, JPEG compression noise, spatially varying illumination, local defocus blur, and random erasing. These augmentations increase the diversity of background artifacts and stroke appearances, improving robustness to the unseen degradation patterns encountered in the leave-one-year-out setting.

\subsection{Leave-One-Year-Out Evaluation Protocol}
To evaluate robustness under unseen document degradation domains, we adopt a full leave-one-year-out (LOO) protocol across DIBCO-style benchmark years. In each fold, one benchmark year is held out as the unseen target domain, while all remaining years are used for training and validation. All images from the held-out target year are completely excluded from both training and validation, and are used only for final testing. This process is repeated for every year. Formally, given a set of benchmark years $\mathcal{Y}$ and a held-out target year $y_t$, the split is defined as:
\begin{equation}
    \mathcal{D}_{test} = \mathcal{D}_{y_t}, \quad
    \mathcal{D}_{train} = \bigcup_{y \in \mathcal{Y}, y \neq y_t} \mathcal{D}_{y}.
\end{equation}

This protocol is more challenging than random image-level splitting because no sample from the target year is used during training or validation. Since different DIBCO/H-DIBCO years contain different acquisition conditions, document types, and degradation patterns, each held-out year is treated as an unseen degradation domain. This setting evaluates whether DR-Mamba's forward-pass, sample-conditioned adaptation generalizes to unseen degradation domains relative to non-adaptive single-route modeling. We additionally highlight the 2019 fold because it contains severe background interference and low-contrast strokes, and use it as the unified setting for detailed ablations.

\subsection{Evaluation Metrics}
Following standard DIBCO protocols, we report F-measure (FM), pseudo-F-measure (p-FM), peak signal-to-noise ratio (PSNR), and distance reciprocal distortion (DRD). All four metrics are computed with the official DIBCO/H-DIBCO evaluation tools rather than a custom re-implementation, so our scores follow exactly the same definitions and weighting as the original contest protocols. FM scores foreground-background classification from precision and recall; p-FM emphasizes skeletonized thin-stroke preservation; PSNR measures pixel-level similarity to the ground truth; and DRD reflects perceptual distortion from binarization errors. Higher FM, p-FM, and PSNR and lower DRD indicate better performance.

\subsection{Comparison with Existing Methods}
Table~\ref{tab:overall_comparison} compares DR-Mamba with classical and learning-based document binarization methods. Because the competing numbers are collected from prior work under differing train/test year splits and implementation details, this table is a reference comparison rather than a strictly controlled re-implementation. Within this caveat, DR-Mamba attains the highest p-FM among all compared methods---the metric most directly tied to skeletonized thin-stroke preservation---and stays within 0.4 FM of the best-reported result under a single fixed protocol. Some transformer-based methods report lower DRD, reflecting very smooth boundaries; we instead prioritize faint- and broken-stroke recovery, which is reflected in our p-FM and, as shown next, in robustness on the hardest held-out fold. Our main evidence remains the controlled leave-one-year-out evaluation and ablation studies.

\begin{table}[!htbp]
\centering
\caption{Overall comparison with classical and learning-based document image binarization methods. Competing numbers are taken from prior work under differing train/test year splits and implementation details, so this table is a reference comparison rather than a strictly controlled re-implementation. Higher FM, p-FM, and PSNR are better; lower DRD is better. ``Years'' denotes the number of benchmark years in the reported average.}
\label{tab:overall_comparison}
\footnotesize
\begin{tabular}{lccccc}
\toprule
Method & FM$\uparrow$ & p-FM$\uparrow$ & PSNR$\uparrow$ & DRD$\downarrow$ & Years \\
\midrule
Otsu~\cite{otsu1979threshold}              & 77.27 & 79.14 & 15.22 & 17.24 & 10 \\
Sauvola~\cite{sauvola2000adaptive}         & 78.10 & 82.91 & 15.77 & 8.72 & 10 \\
Gatos~\cite{gatos2006adaptive}             & 87.33 & 89.85 & 17.98 & 5.40 & 10 \\
RDD~\cite{su2013robust}                    & 85.96 & 87.43 & 17.92 & 7.00 & 10 \\
DeepOtsu~\cite{he2019deepotsu}             & 88.59 & 90.84 & 19.33 & 3.76 & 10 \\
DD-GAN~\cite{deng2020document}             & 88.98 & 91.17 & 19.87 & 3.57 & 10 \\
cGANs~\cite{zhao2019cascaded}              & 90.24 & 91.97 & 19.66 & 3.81 & 10 \\
DocBinFormer~\cite{biswas2023docbinformer} & 92.22 & 94.09 & 21.30 & \textbf{0.15} & 10 \\
D$^2$BFormer~\cite{yang2023d2bformer}      & 91.75 & 93.05 & 20.85 & 2.71 & 10 \\
TransDocUNet~\cite{sukesh2023transdocunet} & \textbf{92.44} & 93.86 & \textbf{22.16} & 0.49 & 10 \\
\midrule
DR-Mamba                                   & 92.06 & \textbf{94.15} & 20.91 & 2.26 & 10 \\
\bottomrule
\end{tabular}
\end{table}

\subsection{Leave-One-Year-Out Results}
Table~\ref{tab:loo_results} reports the full leave-one-year-out results of DR-Mamba. The model achieves an average FM of 92.06 and p-FM of 94.15 across ten held-out years, indicating stable thin-stroke preservation under unseen degradation domains. Performance remains high on most years, while the 2019 fold is clearly the most difficult split.

\begin{table}[!htbp]
\centering
\caption{Leave-one-year-out evaluation results across DIBCO/H-DIBCO benchmark years. Higher FM, p-FM, PSNR are better; lower DRD is better.}
\label{tab:loo_results}
\footnotesize
\begin{tabular}{lcccc}
\hline
Target Year & FM $\uparrow$ & p-FM $\uparrow$ & PSNR $\uparrow$ & DRD $\downarrow$ \\
\hline
2009 & 94.12 & 96.24 & 21.33 & 1.3927 \\
2010 & 93.74 & 96.10 & 21.45 & 1.3125 \\
2011 & 94.96 & 97.24 & 21.73 & 1.3463 \\
2012 & 97.20 & 97.96 & 23.79 & 0.9698 \\
2013 & 96.85 & 98.07 & 23.01 & 1.1766 \\
2014 & 97.33 & 98.81 & 23.37 & 0.7332 \\
2016 & 90.45 & 93.91 & 19.84 & 3.4212 \\
2017 & 90.94 & 93.73 & 18.97 & 2.9187 \\
2018 & 89.87 & 93.32 & 19.66 & 2.7691 \\
2019 & 75.16 & 76.08 & 15.92 & 6.5239 \\
\hline
Average & 92.06 & 94.15 & 20.91 & 2.2564 \\
\hline
\end{tabular}
\end{table}

Among all held-out folds, the 2019 split is the most challenging due to severe background interference and low-contrast foreground strokes. This is reflected by its lower FM, p-FM, and PSNR and higher DRD compared with the other years. We therefore further compare DR-Mamba with existing methods on the 2019 fold in Table~\ref{tab:comparison_2019}.

\begin{table}[!htbp]
\centering
\caption{Comparison on the challenging 2019 held-out fold. Higher FM, p-FM, and PSNR are better. The competing results are taken from their original papers as external reference numbers and may follow different training protocols; they are not strictly comparable to our leave-one-year-out setting.}
\label{tab:comparison_2019}
\footnotesize
\begin{tabular}{lccc}
\toprule
Method & FM$\uparrow$ & p-FM$\uparrow$ & PSNR$\uparrow$ \\
\midrule
Otsu~\cite{otsu1979threshold}              & 63.87 & 60.24 & 12.67 \\
Sauvola~\cite{sauvola2000adaptive}         & 63.82 & 60.18 & 12.66 \\
DeepOtsu~\cite{he2019deepotsu}             & 60.75 & 59.91 & 14.44 \\
DD-GAN~\cite{deng2020document}             & 57.96 & 57.30 & 14.43 \\
DocEnTr~\cite{souibgui2022docentr}         & 59.00 & 60.00 & 13.85 \\
DocDiff~\cite{docdiff}                     & 73.38 & 75.12 & 15.14 \\
DocBinFormer~\cite{biswas2023docbinformer} & 60.31 & 64.00 & 14.49 \\
TransDocUNet~\cite{sukesh2023transdocunet} & 56.79 & 56.78 & 13.36 \\
D$^2$BFormer~\cite{yang2023d2bformer}      & 67.63 & 66.69 & 15.05 \\
MFE-GAN~\cite{mfegan}                      & 70.41 & 70.96 & 13.79 \\
NAF-DPM~\cite{nafdpm}                      & 74.61 & \textbf{76.25} & 15.39 \\
\midrule
DR-Mamba                                   & \textbf{75.16} & 76.08 & \textbf{15.92} \\
\bottomrule
\end{tabular}
\end{table}

As shown in Table~\ref{tab:comparison_2019}, DR-Mamba achieves the best FM and PSNR on the challenging 2019 fold, with p-FM only 0.17 behind NAF-DPM. Notably, several methods with strong aggregate scores in Table~\ref{tab:overall_comparison}---including DocBinFormer and TransDocUNet---collapse here (FM below 61), whereas DR-Mamba degrades far more gracefully. This indicates that strong average performance does not guarantee robustness on the most severely degraded fold, where adaptive detail-background suppression is most valuable. These external numbers serve only as supporting evidence.

\begin{figure}[t]
\centering
\setlength{\tabcolsep}{2pt}
\renewcommand{\arraystretch}{0.8}
\begin{tabular}{cccc}
\textbf{Input} & \textbf{GT} & \textbf{w/o DR-Mamba} & \textbf{DR-Mamba} \\
\includegraphics[width=0.235\linewidth]{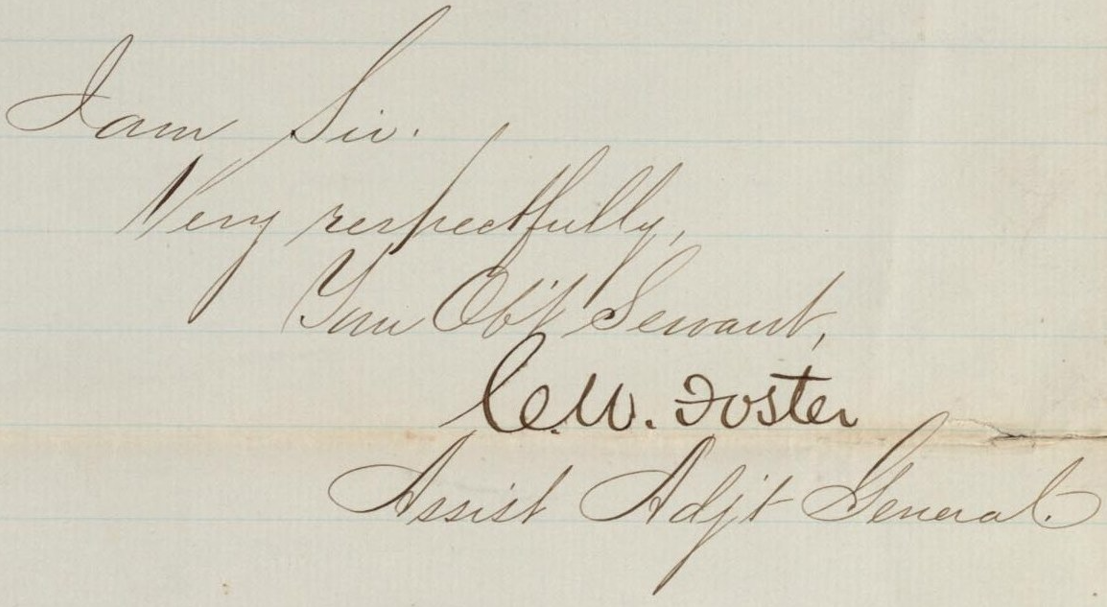} &
\includegraphics[width=0.235\linewidth]{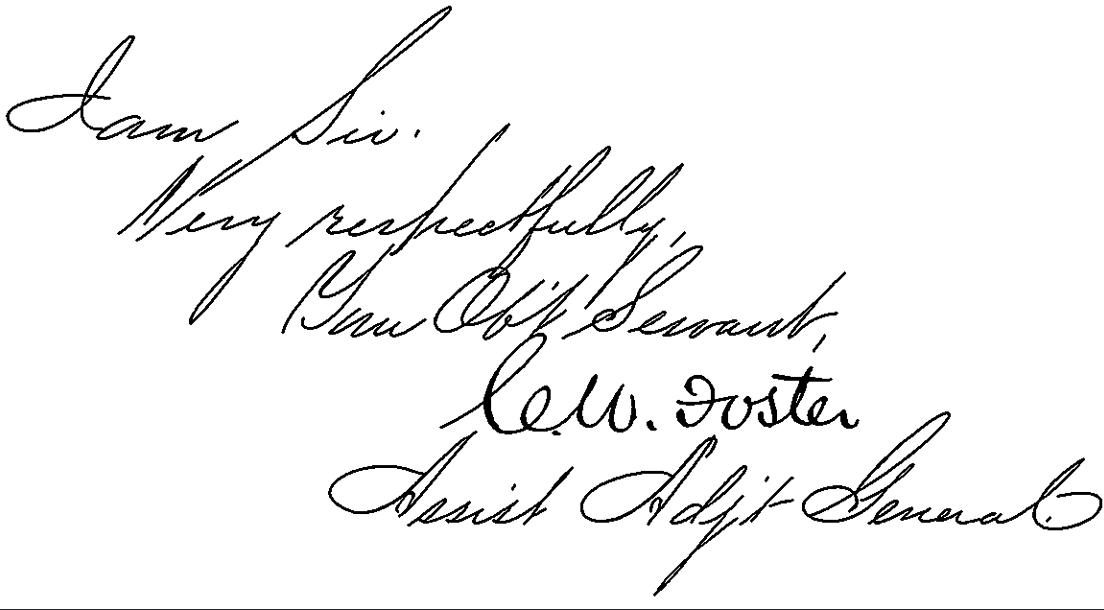} &
\includegraphics[width=0.235\linewidth]{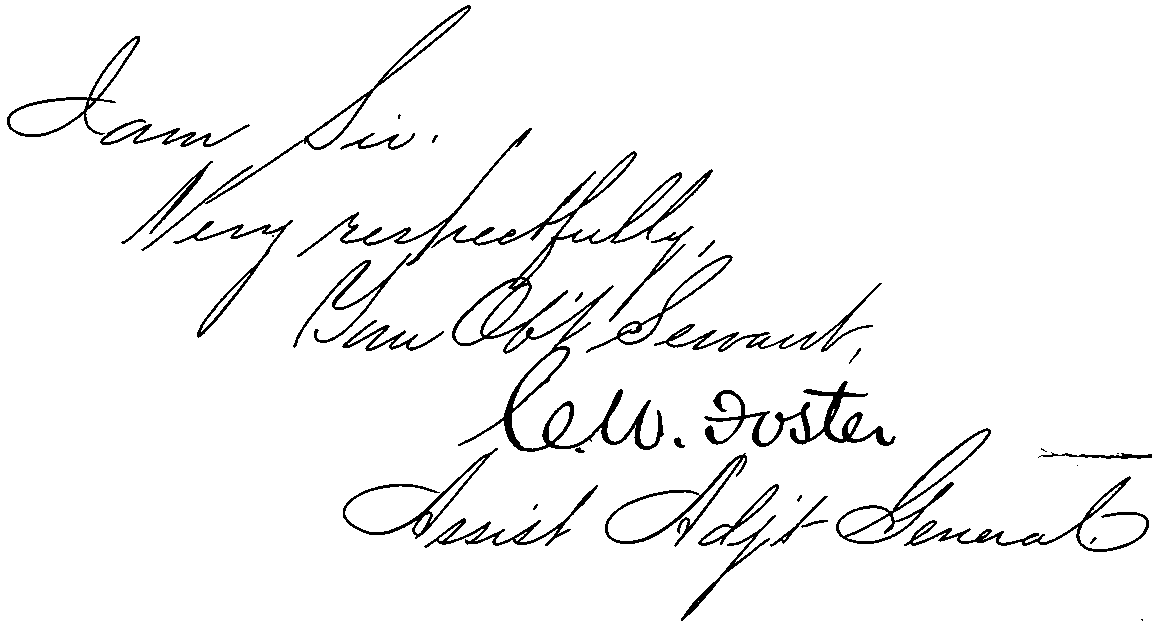} &
\includegraphics[width=0.235\linewidth]{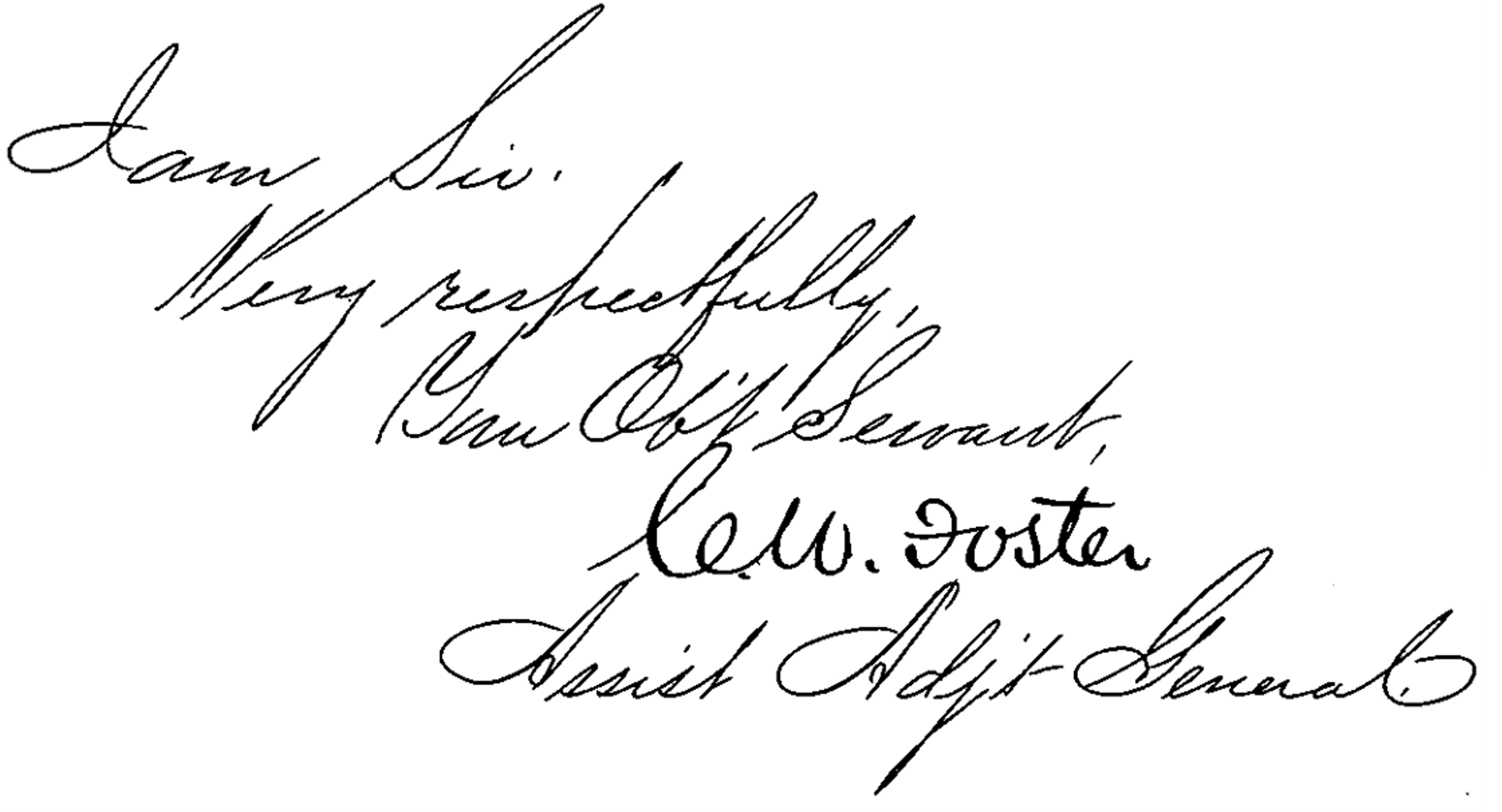} \\

\includegraphics[width=0.235\linewidth]{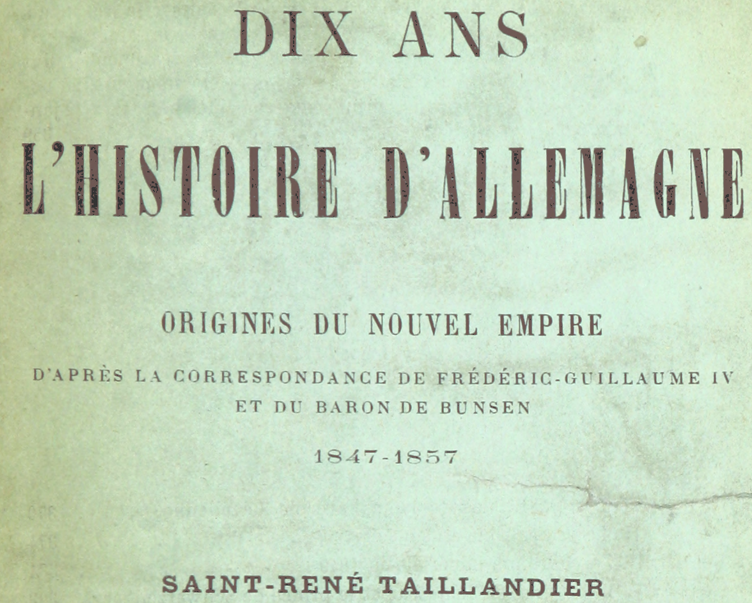} &
\includegraphics[width=0.235\linewidth]{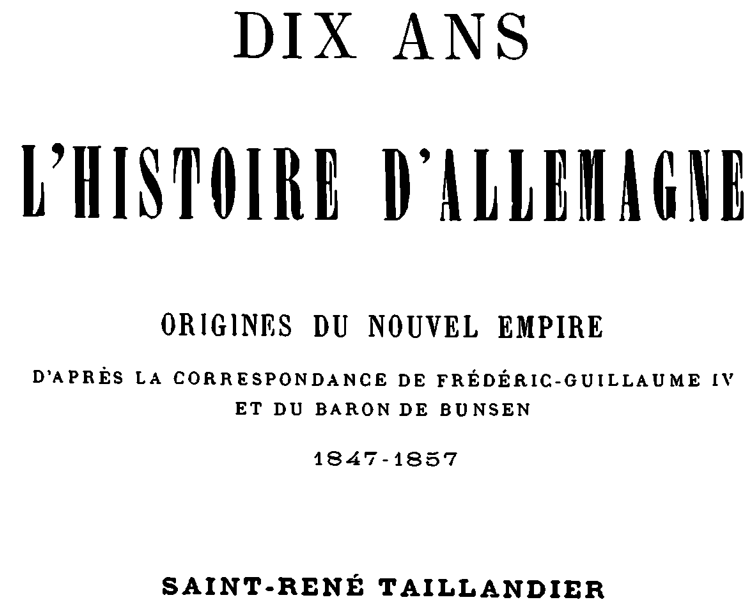} &
\includegraphics[width=0.235\linewidth]{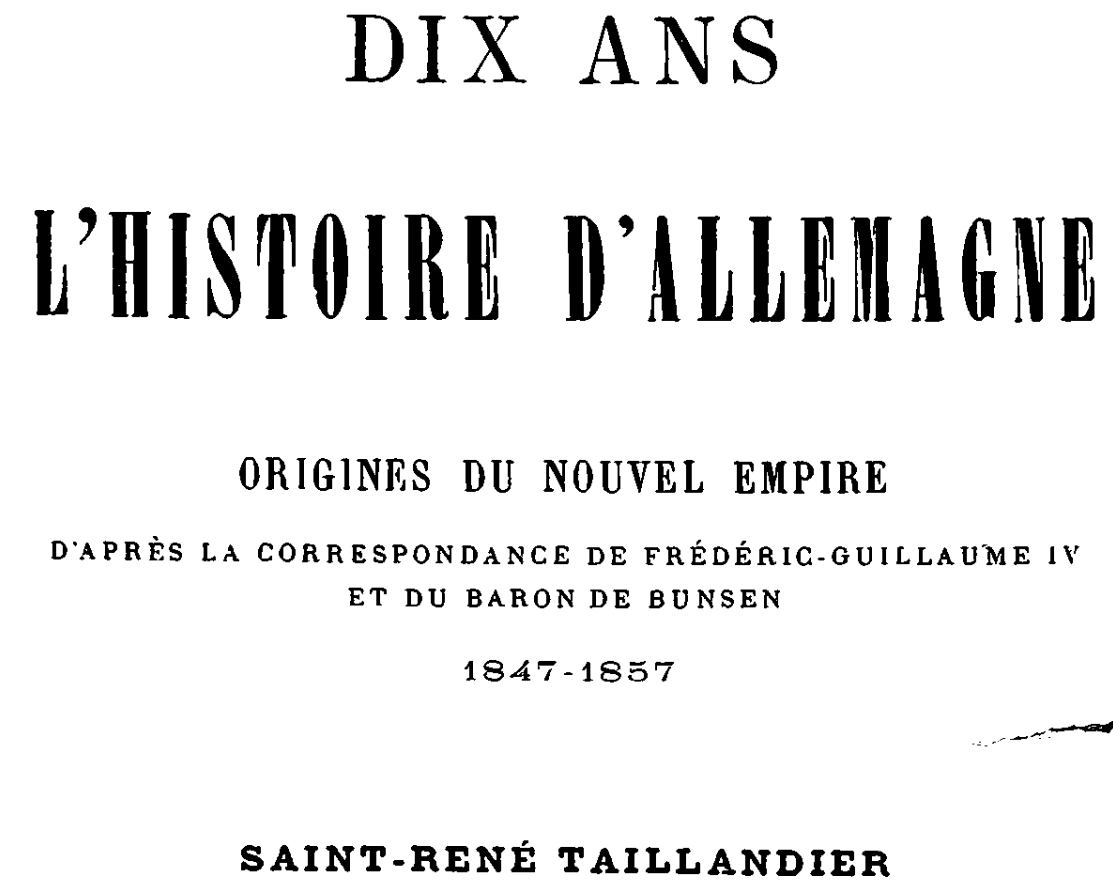} &
\includegraphics[width=0.235\linewidth]{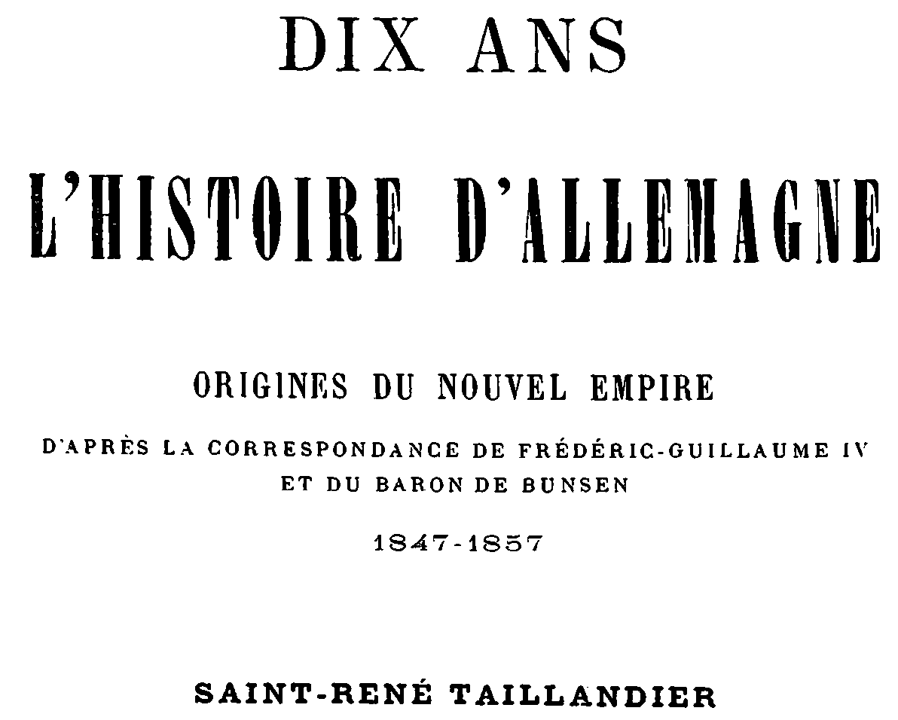} \\

\includegraphics[width=0.235\linewidth]{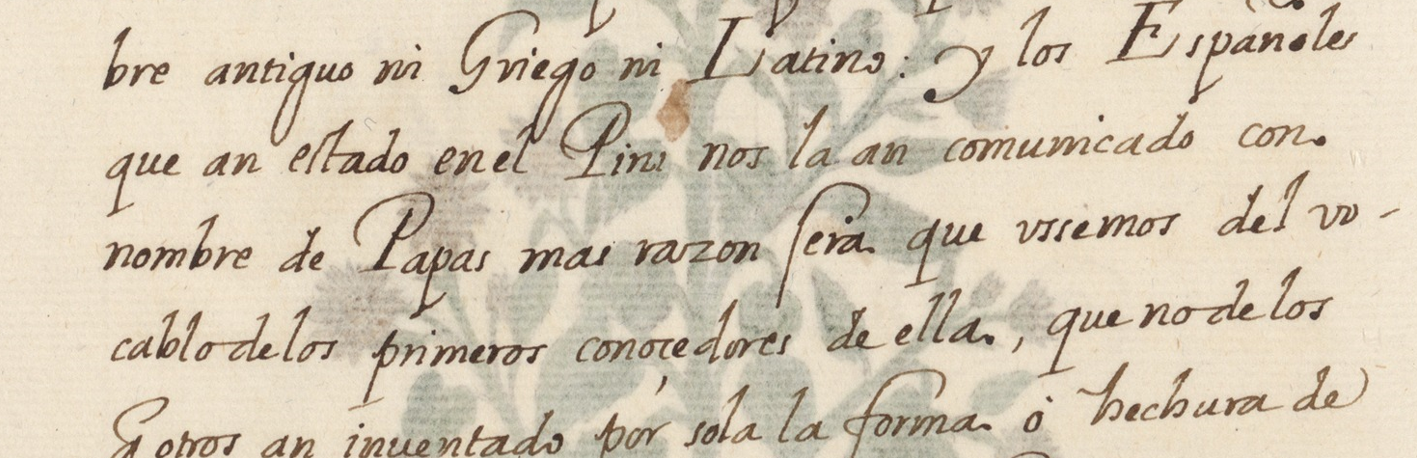} &
\includegraphics[width=0.235\linewidth]{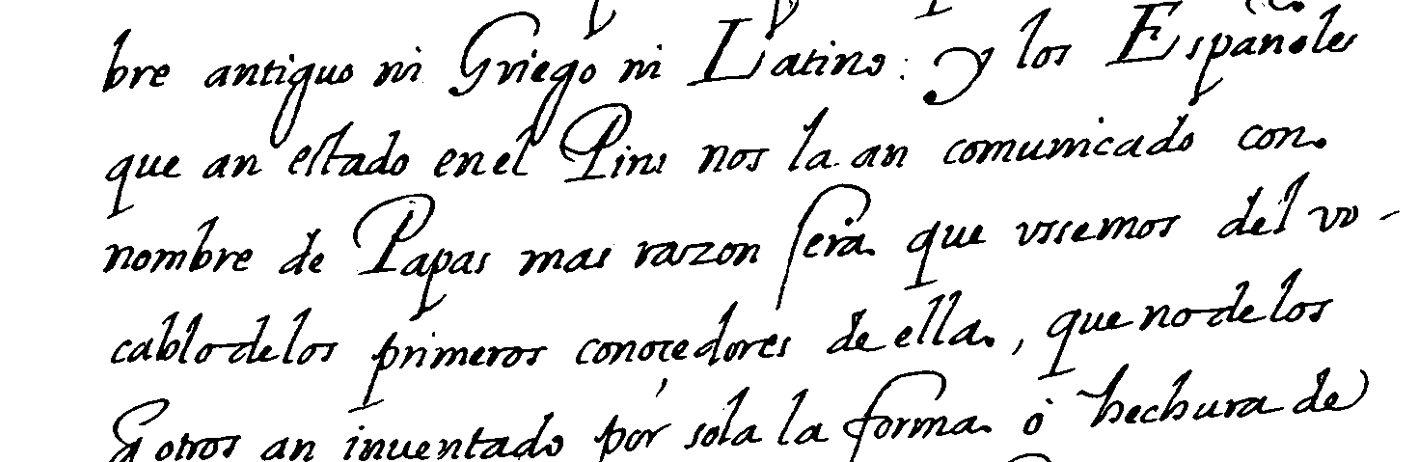} &
\includegraphics[width=0.235\linewidth]{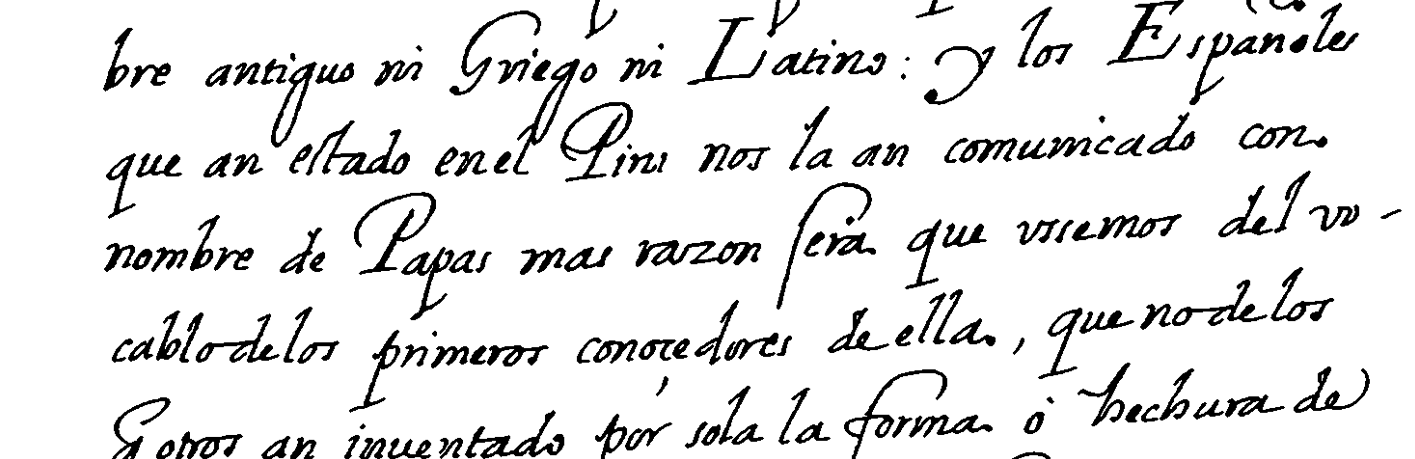} &
\includegraphics[width=0.235\linewidth]{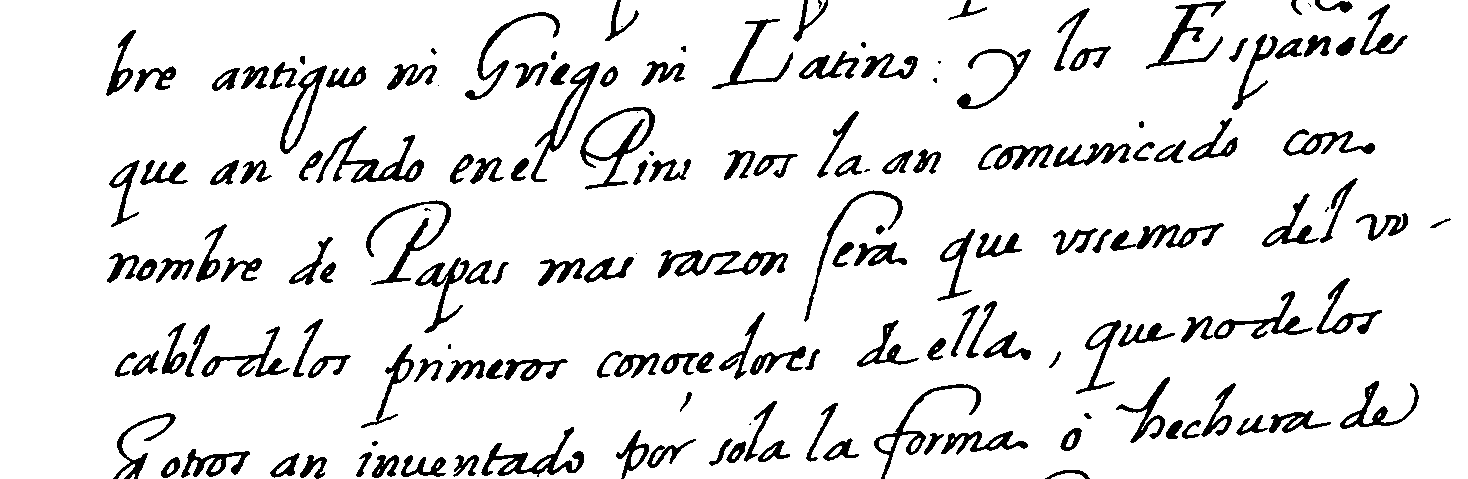} \\
\end{tabular}
\caption{Qualitative comparison on challenging held-out documents. DR-Mamba better suppresses background degradation while preserving thin strokes and stroke continuity.}
\label{fig:qualitative_comparison}
\end{figure}

Figure~\ref{fig:qualitative_comparison} further provides qualitative examples on challenging held-out documents.

\subsection{Ablation Study}
\label{sec:ablation}
We conduct ablation studies to verify the contribution of each component. We first report average results across all leave-one-year-out folds to test whether the gains are consistent across unseen domains. We then perform detailed component-wise and design-specific ablations on the challenging 2019 fold.

\begin{table}[!htbp]
\centering
\caption{Average ablation results across all leave-one-year-out folds. Each row progressively adds one component, following the 2019-fold incremental study in Table~\ref{tab:ablation_main}.}
\label{tab:ablation_average}
\footnotesize
\begin{tabular}{llcccc}
\toprule
ID & Model & Avg FM$\uparrow$ & Avg p-FM$\uparrow$ & Avg PSNR$\uparrow$ & Avg DRD$\downarrow$ \\
\midrule
A0 & Baseline (single-route Mamba, BCE only)        & 88.74 & 90.91 & 19.27 & 4.1108 \\
A1 & + Dual route (concatenation fusion)            & 89.31 & 91.48 & 19.55 & 3.7642 \\
A2 & + Adaptive suppression ($D-\beta B$)           & 90.83 & 92.86 & 20.14 & 3.0426 \\
A3 & + Detail-guided reconstruction                 & 91.58 & 93.49 & 20.55 & 2.6189 \\
A4 & + Thin-stroke-aware loss (full DR-Mamba)       & 92.06 & 94.15 & 20.91 & 2.2564 \\
\bottomrule
\end{tabular}
\end{table}

\subsubsection{Component-wise Incremental Ablation.}
Table~\ref{tab:ablation_main} reports the incremental ablation on the 2019 fold. The baseline (A0) is a ConvNeXt-Tiny encoder-decoder followed by a conventional single-route Mamba block, supervised only by binary cross-entropy. We then progressively build the proposed framework by adding one design component at a time: a dual-route decomposition with naive concatenation fusion (A1), the proposed adaptive suppression that replaces concatenation with $D-\beta B$ (A2), the detail-guided reconstruction module (A3), and finally the thin-stroke-aware compound loss (A4), which recovers the full DR-Mamba model. This incremental study shows how the complete framework is progressively constructed from the baseline. To avoid attributing all gains to a single factor, we further isolate the fusion design and the loss design in the following focused ablation studies.

\begin{table}[!htbp]
\centering
\caption{Component-wise incremental ablation on the 2019 fold. Each row progressively adds one component to the previous configuration. A0 is the single-route baseline and A4 is the full DR-Mamba.}
\label{tab:ablation_main}
\footnotesize
\begin{tabular}{llcccc}
\hline
ID & Configuration & FM $\uparrow$ & p-FM $\uparrow$ & PSNR $\uparrow$ & DRD $\downarrow$ \\
\hline
A0 & Baseline (single-route Mamba, BCE only) & 69.30 & 70.55 & 13.40 & 9.7012 \\
A1 & + Dual route (concatenation fusion) & 70.21 & 71.67 & 13.97 & 9.1026 \\
A2 & + Adaptive suppression ($D-\beta B$) & 72.96 & 74.32 & 14.65 & 8.5742 \\
A3 & + Detail-guided reconstruction & 74.41 & 75.05 & 15.31 & 7.7415 \\
A4 & + Thin-stroke-aware loss (full DR-Mamba) & 75.16 & 76.08 & 15.92 & 6.5239 \\
\hline
\end{tabular}
\end{table}

\subsubsection{Effect of the Adaptive Suppression Fusion.}
Table~\ref{tab:ablation_fusion} isolates the fusion design by fixing the dual-route decomposition and varying only the route fusion operation. Adaptive subtraction outperforms addition, concatenation, and fixed-$\beta$ subtraction, supporting our claim that binarization benefits from input-dependent background suppression rather than generic feature fusion. Removing the fast--slow decay gap ($A_D{=}A_B$) while keeping the subtractive gate degrades all metrics, confirming that the asymmetric decay between the two routes---not merely the dual-route structure---is responsible for the gain.

\begin{table}[!htbp]
\centering
\caption{Effect of the fusion operation between the detail route $D$ and the background route $B$, with the dual-route decomposition fixed.}
\label{tab:ablation_fusion}
\footnotesize
\begin{tabular}{lcccc}
\hline
Fusion Operation & FM $\uparrow$ & p-FM $\uparrow$ & PSNR $\uparrow$ & DRD $\downarrow$ \\
\hline
Additive $D+B$ & 68.43 & 69.82 & 13.21 & 9.8471 \\
Concatenation $[D,B]$ & 70.21 & 71.67 & 13.97 & 9.1026 \\
Subtraction $D-\beta B$ ($\beta$ fixed) & 71.93 & 73.10 & 14.28 & 8.8305 \\
Subtraction $D-\beta B$ ($\beta$ input-dependent) & 72.96 & 74.32 & 14.65 & 8.5742 \\
\hline
Subtraction $D-\beta B$ (shared decay, $A_D{=}A_B$) & 71.16 & 72.73 & 14.15 & 8.9403 \\
\hline
\end{tabular}
\end{table}


\subsubsection{Effect of the Number of Scanning Directions.}
The DR-Mamba block aggregates selective scans along multiple orientations to capture long-range stroke continuity. Table~\ref{tab:ablation_scan} studies how the number of scanning directions affects performance, comparing a single horizontal scan, a four-directional scan, and an eight-directional scan. We observe that the four-directional scan offers the best trade-off: a single direction is insufficient to model stroke continuity in arbitrary orientations, whereas increasing to eight directions does not yield further gains. We attribute the slight regression at eight directions to the additional diagonal scans aggregating low-contrast background fibers along oblique orientations, which competes with the subtractive suppression mechanism; this is consistent with the predominantly horizontal-vertical stroke orientations in DIBCO documents. We therefore adopt four-directional scanning as the default configuration.

\subsubsection{Effect of the Thin-Stroke-Aware Compound Loss.}
The compound loss aggregates six terms with complementary roles. To verify that the gain does not come from a single dominant term, Table~\ref{tab:ablation_loss} decomposes the loss into groups added incrementally on top of BCE: the boundary and balance terms ($\mathcal{L}_{sdf}$, $\mathcal{L}_{bnd}$, $\mathcal{L}_{tv}$) and the thin-stroke connectivity terms ($\mathcal{L}_{pfm}$, $\mathcal{L}_{cldice}$). The results indicate that the boundary and balance terms mainly sharpen foreground-background boundaries and reduce DRD, whereas the thin-stroke connectivity terms primarily improve p-FM by preserving fine and broken strokes. The two groups are complementary, and the full compound loss achieves the best overall result.

\begin{table}[!htbp]
\centering
\caption{Effect of the number of selective scanning directions in the DR-Mamba block on the 2019 fold.}
\label{tab:ablation_scan}
\footnotesize
\begin{tabular}{lcccc}
\hline
Scanning Directions & FM $\uparrow$ & p-FM $\uparrow$ & PSNR $\uparrow$ & DRD $\downarrow$ \\
\hline
1 direction & 72.94 & 74.40 & 15.18 & 7.4521 \\
4 directions (default) & 75.16 & 76.08 & 15.92 & 6.5239 \\
8 directions & 74.27 & 75.67 & 15.78 & 6.7498 \\
\hline
\end{tabular}
\end{table}

\begin{table}[!htbp]
\centering
\caption{Decomposition of the thin-stroke-aware compound loss on the 2019 fold. Terms are added incrementally on top of BCE.}
\label{tab:ablation_loss}
\footnotesize
\begin{tabular}{lcccc}
\hline
Loss Configuration & FM $\uparrow$ & p-FM $\uparrow$ & PSNR $\uparrow$ & DRD $\downarrow$ \\
\hline
BCE only & 74.41 & 75.05 & 15.31 & 7.7415 \\
+ Boundary \& balance ($\mathcal{L}_{sdf}, \mathcal{L}_{bnd}, \mathcal{L}_{tv}$) & 74.79 & 75.34 & 15.68 & 6.8013 \\
+ Thin-stroke connectivity ($\mathcal{L}_{pfm}, \mathcal{L}_{cldice}$) & 75.16 & 76.08 & 15.92 & 6.5239 \\
\hline
\end{tabular}
\end{table}

\noindent
Overall, the ablations confirm that the adaptive suppression $D-\beta B$ is the key contributor and that the asymmetric fast--slow decay (not merely the dual-route structure) drives the gain, while four-directional scanning, detail-guided reconstruction, and the thin-stroke-aware loss add complementary improvements in boundary sharpness and thin-stroke connectivity.

\subsection{Backbone Study}
The previous experiments ablate the components of the proposed DR-Mamba block while keeping the backbone fixed. Here we separately study the choice of the backbone encoder, since this is a design decision about feature extraction rather than a component of our contribution. Table~\ref{tab:backbone} compares a VMamba backbone~\cite{liu2024vmamba} with the ConvNeXt-Tiny backbone, with the DR-Mamba block unchanged in both cases. We find that ConvNeXt-Tiny provides stronger hierarchical features for dense binarization prediction under the cross-domain LOO setting. We note that this comparison concerns only the backbone used for feature extraction; the long-range modeling in our framework is still performed by the Mamba-based DR-Mamba block regardless of the backbone choice.

\begin{table}[!htbp]
\centering
\caption{Backbone study on the 2019 fold. The DR-Mamba block is kept unchanged; only the encoder backbone is varied.}
\label{tab:backbone}
\footnotesize
\begin{tabular}{lcccc}
\hline
Backbone & FM $\uparrow$ & p-FM $\uparrow$ & PSNR $\uparrow$ & DRD $\downarrow$ \\
\hline
VMamba & 72.36 & 75.09 & 14.51 & 7.8413 \\
ConvNeXt-Tiny (default) & 75.16 & 76.08 & 15.92 & 6.5239 \\
\hline
\end{tabular}
\end{table}

\section{Conclusion}
We presented DR-Mamba, a sample-conditioned detail-background suppression framework that performs automatic inference-time domain adaptation for degraded document image binarization. The key idea is to adapt Mamba-style selective scanning to the suppression nature of binarization through fast--slow detail-background routes and input-dependent subtractive fusion, $D-\beta B$, where the input image itself acts as the adaptation signal that reconfigures the suppression gate $\beta$ for each unseen document. Full-resolution detail-guided reconstruction and thin-stroke-aware supervision further improve stroke continuity and boundary quality. Under leave-one-year-out evaluation on DIBCO-style benchmarks, where each held-out year is an unseen degradation domain, DR-Mamba shows competitive cross-domain performance and strong results on the most challenging 2019 fold; ablations confirm that adaptive subtractive suppression is the dominant contributor. DR-Mamba thus offers a lightweight, label-free, forward-pass form of automatic domain adaptation for document analysis. Future work will combine this inference-time sample-conditioning with explicit source-free test-time adaptation to further close the gap on the most severely degraded domains. \\

\noindent\textbf{Acknowledgments}\\
This research work is partially supported by National Science and Technology Council, Taiwan, under grant number: 114-2221-E-032-011-.

\bibliographystyle{splncs04}
\bibliography{references}

\end{document}